\documentclass[sigconf, nonacm]{acmart}

\AtBeginDocument{%
  }

\def\code#1{\texttt{#1}}

\usepackage{enumitem}
\usepackage{array}

\usepackage{listings}
\usepackage{xcolor}
\usepackage{stfloats}
\usepackage{float}

\begin{document}

\title{Towards Code-Oriented LM Embeddings for Surrogate-Assisted Neural Architecture Search}

\author{Pranav Somu}
\email{psomu3@gatech.edu}
\orcid{0009-0008-8633-9705}
\affiliation{%
  \institution{Georgia Institute of Technology}
  \city{Atlanta}
  \state{Georgia}
  \country{USA}
}

\author{Advay Balakrishnan}
\email{abb32@gatech.edu}
\orcid{0009-0001-2653-2596}
\affiliation{%
  \institution{Georgia Institute of Technology}
  \city{Atlanta}
  \state{Georgia}
  \country{USA}
}

\author{Stepan Kravtsov}
\email{skravtsov3@gatech.edu}
\orcid{0009-0003-8639-7589}
\affiliation{%
  \institution{Georgia Institute of Technology}
  \city{Atlanta}
  \state{Georgia}
  \country{USA}
}

\author{Aaron McDaniel}
\authornote{Both authors contributed equally to this research as senior authors.}
\email{Aaron.Mcdaniel@gtri.gatech.edu}
\orcid{0000-0002-8387-0444}
\affiliation{
  \institution{Georgia Tech Research Institute}
  \city{Atlanta}
  \state{Georgia}
  \country{USA}
}

\author{Jason Zutty}
\authornotemark[1]
\email{Jason.Zutty@gtri.gatech.edu}
\orcid{0000-0001-7977-1454}
\affiliation{
  \institution{Georgia Tech Research Institute}
  \city{Atlanta}
  \state{Georgia}
  \country{USA}
}

\begin{abstract}
Developing effective surrogates (performance predictors) for Neural Architecture Search (NAS) typically requires expensive fine-tuning or the engineering of complex representations. We propose a low-cost embedding strategy that leverages the inductive bias of Language Models (LMs) to eliminate these overheads. By representing architectures as PyTorch class definition text, we demonstrate that off-the-shelf LMs act as competitive feature extractors without NAS-specialized fine-tuning. The final predictor is constructed by passing the extracted Code-Oriented LM Embeddings (COLE) through a lightweight regression head. We also investigate strategies to improve embedding quality and utilization. Our experiments on the NAS-Bench-201 and einspace search spaces reveal that raw code inputs yield higher predictive performance than other text-based encodings (e.g., ONNX-to-text encodings) when using frozen LMs. We also observe COLE drives superior surrogate-assisted search using the BANANAS algorithm in NAS-Bench-201. When optimizing for CIFAR-100 performance, replacing structural path encodings with COLE for architecture representation allows for a 34\% decrease in the evaluation budget required to reach within 1\% of the fittest architecture in the search space (by test accuracy). As any neural architecture can be represented as code, these findings establish COLE as a versatile and efficient foundation for advancing NAS.\footnote{This is an extended version of work accepted to GECCO 2026: \url{https://doi.org/10.1145/3795101.3805435}}\footnote{In the official print, we reported the CIFAR-100 evaluation metric as `validation accuracy,' but we correct the terminology to `test accuracy' in this print to accurately reflect the evaluation data provided by the NAS library we used.}\footnote{Our code is available at \url{https://github.com/pcsom/cole/tree/v1.0}}
\end{abstract}




\maketitle

\section{Introduction}
Neural Architecture Search (NAS) automates the discovery of high-performing neural architectures for a given task in a search space. However, conducting NAS in large search spaces becomes computationally expensive due to complex evaluation functions. While the objective for identifying an optimal architecture in a search space is simple to formulate (e.g., minimizing a loss metric), the cost of evaluating the fitness function renders brute force searches computationally intractable. This computational bottleneck motivates the area of performance predictors (surrogates), which seek to approximate the fitness function using low-cost estimators ranging from simple Learning Curve Extrapolation \cite{yan2021bench} to complex Graph Neural Networks (GNNs) \cite{akhauri2024encodings}, effectively substituting the expensive ground-truth evaluations with quick inferences. For this method, architectures must be converted into input representations compatible with surrogate models.

Structural representations such as adjacency matrices or path encodings capture network topology sufficiently. However, capturing hyperparameters like convolution kernel sizes requires custom strategies, such as registering all acceptable values as separate configurations, which is often complex to implement or computationally inefficient \cite{qin2025onnx}. Furthermore, these encoding schemes are inherently rigid, typically relying on fixed-dimensional matrices tailored to specific cell-based search spaces. This means applying the predictor to novel architecture spaces requires re-engineering of the encoding scheme to accommodate varying properties like quantity and arrangements of operations. Recent efforts to make generalized encodings have turned to Language Models (LMs), proposing them as universal regressors. Notable frameworks such as in Qin et al. \cite{qin2025transferrable} have attempted to unify NAS by converting architectures into text-based encodings and fine-tuning encoder-based language models (e.g., ModernBERT) to interpret them. Qin et al. \cite{qin2025transferrable} translated einspace \cite{ericsson2024einspace}, an expressive search space which builds architectures from fundamental operations, into derivation tree strings. These strings were used to fine-tune ModernBERT Large end-to-end to predict architecture performance. While effective, the method introduces the high computational cost of fine-tuning the LM to learn a foreign representation format. By requiring a warm-up phase of fine-tuning, such methods counteract the efficiency gains of using a surrogate model. Instead, it would be beneficial for a search algorithm to identify promising regions of the search space out-of-the-box, enabling cold-start searches.

A key insight that drives our work is that neural architectures naturally have a code representation. Language models like CodeLlama and ModernBERT have been pre-trained on trillions of tokens, some (or most) of which are code \cite{grattafiori2024llama, warner2025smarter}. In contrast, text-based encodings like derivation tree strings or ONNX-to-text encoding \cite{qin2025onnx} are underrepresented in training data thus misaligned with the models’ learned distributions. We demonstrate that native code representations utilize the LM's out-of-the-box interpretation capabilities more effectively than other misaligned text-based formats.

In this work, we investigate the efficacy of using frozen LMs as zero-shot embedding engines for NAS. Unlike prior approaches that rely on expensive fine-tuning and text-based encodings, we feed raw PyTorch class definition text directly into frozen LMs and use the model’s hidden state activations to compute a Code-Oriented LM Embedding (COLE). Code can represent any neural architecture, so our work can be incorporated into most surrogate-enabled NAS pipelines. Our contributions are as follows:

\begin{enumerate}[topsep=5pt, itemsep=2pt, parsep=2pt, leftmargin=*]
\item We show that COLE is an embedding strategy superior to other text-based encodings with frozen LMs. We demonstrate that computationally expensive fine-tuning such as in Qin et al. \cite{qin2025transferrable} may not be strictly necessary to achieve effective NAS embeddings.
\item We demonstrate that integrating COLE as a neural architecture representation strategy into an existing surrogate-assisted NAS (SA-NAS) algorithm leads to improved performance compared to representing architectures using structural encodings like path encodings.
\item We document effective tuning applicable to the COLE and regression pipeline, including PyTorch code verbosity and Principal Component Analysis (PCA) for dimensionality reduction.
\end{enumerate}

By validating that frozen LMs, coupled with code as input, can serve as neural architecture embedders out of the box, we pave the way for a search paradigm that is not only search-space agnostic but also removes the expensive pre-training that is required to learn specialized formats of architectures.
\section{Related Work}

\subsection{Search Spaces and NAS Benchmarks}
Search spaces are commonly defined by a cell-based approach, where the cell is a repeatable unit (a block of operations) stacked in groups to form a large architecture \cite{zoph2018learning, ying2019bench, dong2020bench, zela2020surrogate, liu2018darts}. In a cell-based approach, the search algorithm optimizes the cell rather than the entire network topology. Each cell is a directed acyclic graph (DAG) with a predefined operation set between nodes. For example, cells of the NAS-Bench-201 search space \cite{dong2020bench} contain zeroize, skip-connect, convolution, and average pooling operations. More recently, there has been work on defining search spaces based on context-free grammars \cite{ericsson2024einspace, schrodi2023construction}. Einspace, for example, leverages a functional grammar to use fundamental operations to construct architectures, represented as derivation trees. This allows discovery of novel primitives that flexibly derive and combine elements of CNNs, Transformers and MLPs. 
To facilitate reproducible comparison of surrogate models or NAS algorithms, there are widely adopted NAS benchmarks \cite{ying2019bench, dong2020bench, zela2020surrogate}. These benchmarks provide pre-computed architecture-accuracy pairs within their own defined search spaces, allowing easy evaluation during a NAS algorithm execution. We conduct our evaluations using NASLib \cite{white2021powerful}, a modular open-source framework that integrates these benchmarks and provides a unified interface for search strategies. Specifically, we integrate with the BANANAS search algorithm \cite{white2021bananas}, a surrogate-assisted approach, provided by NASLib. BANANAS combines a neural predictor with Bayesian Optimization, generating candidate architectures via mutation and selecting the most promising ones for evaluation based on surrogate predictions. In this work, we integrate COLE directly into the BANANAS pipeline, replacing the structural path encoding, BANANAS' default architecture representation strategy.

\subsection{Graph Predictors and Encodings}
Architectures must be encoded into compatible representations when working with a performance predictor. Early works relied on structural encodings such as adjacency matrices or path encodings \cite{white2021bananas}, which map graph topology to fixed-dimensional spaces. Later, Graph Neural Networks (GNNs) were introduced to learn embeddings directly from variable graph structures. Standard GNNs, however, often treat operations merely as labels; to capture deeper semantics, GATES \cite{ning2020generic} was proposed to explicitly model operation-wise information flow, mimicking the actual data processing of the network. Similarly, SemiNAS \cite{luo2020semi} improved sample efficiency by utilizing pre-training on large unlabeled datasets before fine-tuning. Unsupervised methods such as Arch2Vec \cite{yan2020does} and CATE \cite{yan2021cate} further advanced this direction by using variational autoencoders to map discrete architectures into a continuous latent space. More recently, FLAN \cite{akhauri2024encodings} proposed flow-based graph attention to explicitly capture signal propagation within the architecture, merging the benefits of GNNs and Transformers. HyperNAS \cite{lv2025hypernas} enhanced representation learning by combining a global encoding scheme that employs a Graph Convolutional Network (GCN) to capture cell-level and macro-structural dependencies, while simultaneously leveraging a shared hypernetwork to enhance the generalization and quality of the learned architecture embeddings. 
\subsection{Zero-Cost Proxies}
\subsubsection{Zero-Cost Proxies (ZCPs) Definition.}
\label{sec:subsubsection}
ZCPs are a class of techniques in NAS that estimate performance of a neural network without actually training it \cite{white2021powerful}. Typically, they conduct one forward and backward pass of data to measure gradient flow through the network. ZCPs measure values such as trainability by inspecting gradients or correlations of activations within the network \cite{abdelfattah2021zero}. Some recent work in ZCPs include the Synaptic Flow, Jacobian Covariance, and GradNorm methods \cite{tanaka2020pruning, mellor2021neural, abdelfattah2021zero}.
\subsubsection{Limitations and Recent Advances (RBFleX-NAS)}
\label{sec:subsubsection}
While ZCPs offer extreme efficiency, they have historically struggled with consistency across different search spaces and tasks \cite{abdelfattah2021zero}. A notable recent advancement addressing these limitations is RBFleX-NAS \cite{yamasaki2025rbflex}, which reached a Kendall's Tau value of 0.569 on NAS-Bench-201 for the CIFAR-10 dataset, around a 46\% improvement on average against other ZCPs.
\subsection{LMs in NAS}
Two recent studies by Qin et al. \cite{qin2025transferrable, qin2025onnx} represent the current approaches in LM-based regression surrogates. Both approaches rely on fine-tuning Encoder-only language models (specifically ModernBERT). ONNX-Net \cite{qin2025onnx} includes a separate regression head which intakes the LM embeddings, while the LM in Qin et al. \cite{qin2025transferrable} was modified to train and predict architecture performance directly (end-to-end). Qin et al. \cite{qin2025transferrable} addresses the einspace search space by fine-tuning ModernBERT on ``derivation tree strings,'' the text-based version of einspace's derivation trees that comprise of the context-free grammar. Qin et al. \cite{qin2025transferrable} demonstrated that encoder LMs like ModernBERT can serve as effective predictors in expressive search spaces. Meanwhile, Qin et al. \cite{qin2025onnx} proposed serializing architectures into ONNX-to-text encodings, a universal encoding scheme which can accommodate any search space whose architectures can be converted into ONNX files. 

Other methodologies utilize LMs to actively generate candidate architectures, effectively functioning as the search algorithm itself. LLM Guided Evolution \cite{morris2024llm} employs LLMs to direct mutation and crossover operations. Meanwhile, alternative surrogates utilize few-shot prompting to directly query the LM for performance estimates or fine-tuning the model on extensive datasets of architecture-performance pairs to rank candidates \cite{hu2025lm, jawahar2024llm}.

\subsubsection{Theoretical Foundations for Frozen Embeddings}
\label{sec:subsubsection}
The viability of using frozen LMs for creating embeddings without fine-tuning is supported by recent theoretical analysis from Tang et al. \cite{tang2024understanding}. The work demonstrated that such embeddings exhibit Lipschitz continuity, meaning small semantic edits in the input string result in smooth embedding space transitions. This makes the embeddings act as a stable input space to support lightweight regressors’ (e.g., MLPs) capability to interpolate accuracy values from frozen features. 
Tang et al. \cite{tang2024understanding} also provided empirical evidence regarding Pooling Strategies applied to hidden activations to create an embedding. They demonstrate that on regression tasks, Mean Pooling (averaging representations across the sequence of LM activations) consistently outperforms Last-Token (EOS) pooling, which they found to be non-robust to high-dimensional inputs.

\section{Methodology}

Our approach, COLE, operates on the premise that neural architectures are best represented as executable code when utilizing LMs as processors. This choice leverages the massive pre-training of LMs on software corpora, where code is the native modality.

\subsection{Native Code Stringification}

First, we map the Neural Architecture Search space directly to Python class definitions. We convert architectures into a valid PyTorch-based (\code{torch.nn.Module}) subclass using a deterministic translation.

\subsubsection{Architectures of the NAS-Bench-201 Search Space}\label{code_for_nb201} We illustrate this process using the cell-based NAS-Bench-201 search space. A cell is a unit of operations focused around convolutions. The standard specification for this benchmark uses a pipe-delimited string to represent the graph topology and operations, as shown in Figure~\ref{fig:archstring}. We parse this string to construct a \code{Cell} Python class. Operations within a cell often involve multiple steps, such as the \code{ReLU Activation -> 3x3 2D Convolution -> Batch Normalization} operation. We investigated three levels of verbosity for representing these complex operations in code:

\begin{description}
    \item[Helper Method mode (Figure~\ref{fig:helpermethod})] We define complex operations as separate helper classes (e.g., \code{ReLU\_Conv2d\_BatchNorm}) which are fully implemented and called by the main \code{Cell} definition.
    \item[Inline mode (Figure~\ref{fig:inline})] We define all operation logic within the Cell definition itself using \code{nn.Sequential}. This lists all sub-operations explicitly (e.g., \code{nn.Conv2d}, \code{nn.BatchNorm2d}) without defining external helper classes.
    \item[Excluded Helper Method mode] This is similar to the Helper Method mode, but we do not actually provide the implementation for the helper classes in the code string. The code simply refers to them as if they are imported.

\end{description}

Our primary experiments indicated that all three approaches yielded comparable performances. We primarily use the Excluded Helper Method mode in following experiments due to its token efficiency.

\begin{figure}[t]
  \centering
  \begin{lstlisting}[numbers=none, frame=single]
'|avg_pool_3x3~0|+|nor_conv_1x1~0|skip_connect~1|+
    |nor_conv_1x1~0|skip_connect~1|skip_connect~2|'
  \end{lstlisting}
  
  \caption{Default NAS-Bench-201 string specification representing the graph topology.}
  \label{fig:archstring}
\end{figure}

\begin{figure}[h]
  \begin{lstlisting}[language=Python]
class ReLU_Conv2d_BatchNorm(nn.Module):
  def __init__(self, channels, kernel_size, stride, padding):
    super().__init__()
    self.op = nn.Sequential(
      nn.ReLU(inplace=False),
      nn.Conv2d(channels, channels, kernel_size, stride=stride, padding=padding, bias=False),
      nn.BatchNorm2d(channels)
    )
  def forward(self, x):
    return self.op(x)

class Cell(nn.Module):
  def __init__(self, channels):
    super().__init__()
    self.op_0_1 = nn.AvgPool2d(kernel_size=3, stride=1, padding=1)
    self.op_0_2 = ReLU_Conv2d_BatchNorm(channels, kernel_size=1, stride=1, padding=0)
    self.op_0_3 = ReLU_Conv2d_BatchNorm(channels, kernel_size=1, stride=1, padding=0)
  def forward(self, x):
    node_0 = x
    node_1 = self.op_0_1(node_0)
    node_2 = self.op_0_2(node_0) + node_1
    node_3 = self.op_0_3(node_0) + node_1 + node_2
    return node_3
  \end{lstlisting}
  
  \caption{A sample code representation under the Helper Method mode. Complex operations are fully defined as separate classes (e.g., \code{ReLU\_Conv2d\_BatchNorm}) and instantiated within the \code{Cell}. The Excluded Helper Method mode would simply omit the separate class from the string.}
  \label{fig:helpermethod}
\end{figure}

\begin{figure}[h]
  \begin{lstlisting}[language=Python]
class Cell(nn.Module):
  def __init__(self, channels):
    super().__init__()
    self.op_0_1 = nn.AvgPool2d(kernel_size=3, stride=1, padding=1)
    self.op_0_2 = nn.Sequential(nn.ReLU(inplace=False), nn.Conv2d(channels, channels, kernel_size=1, stride=1, padding=0, bias=False), nn.BatchNorm2d(channels))
    self.op_0_3 = nn.Sequential(nn.ReLU(inplace=False), nn.Conv2d(channels, channels, kernel_size=1, stride=1, padding=0, bias=False), nn.BatchNorm2d(channels))

  def forward(self, x):
    node_0 = x
    node_1 = self.op_0_1(node_0)
    node_2 = self.op_0_2(node_0) + node_1
    node_3 = self.op_0_3(node_0) + node_1 + node_2
    return node_3
  \end{lstlisting}
  
  \caption{The Inline mode code representation. The logic for complex operations are defined inline directly inside the \code{\_\_init\_\_} method using the \code{nn.Sequential} module.}
  \label{fig:inline}
\end{figure}

Beyond the cell structure, the LM’s understanding of the architecture can be influenced by the surrounding context, such as the macro-architecture (backbone) that utilizes the \code{Cell} or the target end-task (e.g., classification). We investigate two context add-ons to investigate effects of input scope:
\begin{description}
    \item[Backbone add-on] We prepend the full PyTorch implementation of the macro-architecture (as a class \code{Network}) to the \code{Cell} definition. We adopt the standard backbone implementation directly from the official NAS-Bench-201 repository. This provides the LM with the complete execution flow but significantly increases the token count.
    \item[Comment add-on] We prepend a natural language docstring summarizing the task (e.g., CIFAR-10 classification), the backbone structure (\code{Cell} stacking setup), and training hyperparameters. This aims to provide high-level details without the overhead of the full backbone.

\end{description}

Our primary experiments indicated that while Backbone Mode achieved best performance, the other approaches are competitive. In following experiments, we omit the add-ons for the sake of token efficiency. Appendix \ref{addon_appendix} shows the content of the context add-ons.

\subsubsection{Architectures of the einspace Search Space}\label{einspaceTranspiler} We also developed a deterministic transpiler that converts the abstract einspace \cite{ericsson2024einspace} derivation tree directly into executable PyTorch code.

The transpiler traverses the architecture's derivation tree starting from the root node. Recursively, for each child node, it identifies the corresponding einspace operation and generates a unique Python variable name (e.g., \code{self.branching\_0}, \code{self.linear\_1}) to ensure valid identifier syntax. The traversal distinguishes between non-terminal and terminal operations.
Non-Terminal Nodes (Composite Modules) represent structural patterns in einspace and are mapped to custom container classes. The encoder recursively generates code for all child nodes first, then passes them as arguments to the container. The descriptions of the non-terminal nodes are as follows:
\begin{description}
    \item[Sequential] Mapped to \code{SequentialModule}, which specifies sequential execution of child modules. Its input arguments consist of a pair or list detailing these modules.
    \item[Branching] Mapped to \code{BranchingModule}, which specifies execution of a branching function (e.g., split/clone the input data), an inner function (a list of parallel sequential branches), and an aggregation function (e.g., concat/add).
    \item[Routing] Mapped to \code{RoutingModule}, which specifies execution of a prerouting function (tensor reshaping), inner function (transformation), and a post routing function (shape restoration).
    \item[Computation] Mapped to \code{ComputationModule}, a wrapper for atomic primitives.
\end{description}
Terminal Nodes (Atomic Primitives) are the leaves of the tree and map directly to specific PyTorch layers or custom helper classes. The transpiler uses regular expressions to parse hyperparameters directly from the operation name string (e.g., parsing \code{128} from \code{linear(128)}).
This traversal builds the final network by concatenating all strings produced by the recursive calls. This effectively creates the initialization method of a typical PyTorch module class. 

See Appendix~\ref{einspace_appendix} for an example of the code and derivation tree representations.

\subsection{Frozen Embedding Extraction}
We utilize pre-trained LMs as fixed feature extractors. We freeze the LM's parameters entirely, avoiding the ``cold start'' computational cost of fine-tuning. Let $h_{t}$ denote the hidden state of the $t$-th token at the last layer of the model, where $t \in \{1, ..., T\}$ and $T$ is the input token sequence length. We perform token-wise mean pooling \cite{tang2024understanding} within the layer to obtain the embedding:
$\frac{1}{T} \sum_{t=1}^{T} h_{t}$.
We default to the CodeLlama Python 7B LM for embedding, as its Python-centric pre-training aligns with COLE's PyTorch syntax, and its moderate size prevents the framework from being computationally prohibitive.

\subsection{The Surrogate Head}
The large embedding dimension (e.g., 4096 for CodeLlama 7B) causes the head to suffer from the curse of dimensionality, especially when in low-data regimes. We apply Principal Component Analysis (PCA) to project the embedding to a smaller dimension vector. Our preliminary experiments showed that reducing to a dimensionality of 128 provides strong performance (though not necessarily optimal).

After dimensionality reduction, the embedding is fed into a lightweight Multi-Layer Perceptron (MLP) to predict the architecture’s accuracy. To balance model size and performance, we use a three-layer MLP (128 hidden units per layer) with LeakyReLU activation and dropout ($p=0.1$). 

While using Mean Squared Error (MSE) as a loss function is standard, ranking is also paramount in NAS. We optimize for Kendall’s Tau ($\tau$) rank correlation by utilizing a Pairwise Hinge Loss \cite{akhauri2024encodings}:
$$L(\hat{y}, y) = \sum_{i,j} \max(0, \epsilon - \text{sign}(y_i - y_j)(\hat{y}_i - \hat{y}_j))$$

$y$ is the ground truth accuracy and $\hat{y}$ is the predicted accuracy. This formulation penalizes ``swapped'' pairs in the ranking order. This approach is highly compatible with rank-dependent selection methods like truncation or tournament selection, widely used in evolutionary optimization or NAS. These selection processes rely on the correct ordering of candidates (e.g., whether architecture A ranks higher than architecture B) rather than their absolute fitness values. By optimizing for Kendall's Tau, which maximizes the likelihood of correctly ordering any given pair, we ensure the surrogate's objective aligns well with the decision mechanism of the search algorithm, increasing the probability that truly superior candidates are selected for reproduction. However, we note that MSE is still useful for absolute fitness prediction, such as in fitness proportionate selection methods.

\section{Experiments}

We validate the COLE framework through various experiments. We first analyze the impact of various components of the framework (e.g., language model, PyTorch code building, or regressor training) on predictive performance. We then compare COLE to other text-based encodings. We finally integrate COLE into the NASLib framework to benchmark search efficiency against standard surrogates on NAS-Bench-201. For a visual analysis of how these different encodings group architectures, see Appendix~\ref{tsne_viz}.

\subsection{Embedding Utilization Analysis}
We utilize the NAS-Bench-201 search space to evaluate the quality of our embeddings. Our objective is to maximize the Kendall’s Tau ($\tau$) rank correlation between predicted scores and ground-truth accuracies. We define a Base configuration and analyze how deviations from this baseline affect performance.

\subsubsection{Experimental Setup}
\label{sec:subsubsection}
We utilized a stratified subsampled cross-validation protocol to ensure statistical significance. The 15,625 architectures in NAS-Bench-201 were partitioned into 10 folds, with each fold being approximately 1,562 in size. In each cross-validation step, one fold served as the validation set, while the remaining 9 folds constituted the training pool. For a given training budget $N$ (e.g., 14, 55, 220), we subsampled $N$ architectures from the entire training pool (across all 9 folds). This subsampling was repeated with 10 random seeds for each fold, resulting in a total of 100 trials per training budget configuration (10 folds $\times$ 10 seeds). Data points were stratified into 5 bins based on ground-truth accuracy to ensure representative sampling. 

Our Base configuration consists of embedding with a frozen CodeLlama Python 7B LM, using the Excluded Helper Method mode for cell verbosity, omitting the Backbone and Comment add-ons, using PCA to reduce embedding dimensionality to 128 components, loading LMs in full parameter precision (no quantization), and training the MLP using a Pairwise Hinge Loss. All following experiments detailed in this study are capable of running on a single NVIDIA H100 GPU.
\begin{table}[t]
  \caption{Average Kendall’s Tau ($\tau$) across different training set sizes when modifying the Base configuration.}
  \label{tab:kendall_results}
  \begin{tabular}{lccccc}
    \toprule
    & \multicolumn{5}{c}{Avg Kendall's Tau at Training Set Size} \\
    \cmidrule(lr){2-6}
    Configuration & 14 & 55 & 220 & 879 & 3516 \\
    \midrule
    Base & 0.463 & 0.547 & 0.620 & 0.703 & \textbf{0.797} \\
    No PCA & \textbf{0.493} & \textbf{0.554} & 0.601 & 0.645 & 0.692 \\
    Inline Mode & 0.474 & 0.545 & 0.617 & 0.701 & \textbf{0.797} \\
    Helper Mode & 0.461 & 0.545 & 0.618 & 0.702 & 0.796 \\
    Backbone Add-on & 0.460 & 0.550 & \textbf{0.630} & \textbf{0.711} & \textbf{0.797} \\
    Comment Add-on & 0.436 & 0.540 & 0.616 & 0.701 & 0.795 \\
    MSE Loss & 0.430 & 0.524 & 0.608 & 0.691 & 0.779 \\
    8-bit quantization & 0.472 & 0.546 & 0.619 & 0.693 & 0.773 \\
    \bottomrule
  \end{tabular}
\end{table}

\begin{table}[t]
  \caption{Average Mean Squared Error (MSE) across different training set sizes using MSE vs Pairwise Hinge loss.}
  \label{tab:mse_results}
  \begin{tabular}{lccccc}
    \toprule
    & \multicolumn{5}{c}{Avg MSE at Training Set Size} \\
    \cmidrule(lr){2-6}
    Loss & 14 & 55 & 220 & 879 & 3516 \\
    \midrule
    Pairwise Loss & 180.98 & 151.17 & 139.09 & 116.24 & 104.21 \\
    MSE Loss & \textbf{180.02} & \textbf{147.83} & \textbf{110.61} & \textbf{71.01} & \textbf{33.74} \\
    \bottomrule
  \end{tabular}
\end{table}

\begin{table}[t]
\caption{Model comparison (Kendall's Tau) across different training set sizes. CL: CodeLlama; MB: ModernBERT}
\label{tab:model_comparison}
  \begin{tabular}{>{\raggedright\arraybackslash}p{2.4cm}ccccc}
    \toprule
    & \multicolumn{5}{c}{Avg Kendall's Tau at Training Set Size} \\
    \cmidrule(lr){2-6}
    LM & 14 & 55 & 220 & 879 & 3516 \\
    \midrule
    CL Python 7B & 0.463 & \textbf{0.547} & 0.620 & 0.703 & 0.797 \\
    CL Python 34B & 0.449 & 0.545 & \textbf{0.625} & \textbf{0.711} & \textbf{0.806} \\
    CL 34B & 0.457 & 0.538 & 0.618 & 0.709 & 0.802 \\
    CL Instruct 34B & 0.463 & 0.539 & 0.620 & 0.709 & 0.801 \\
    Llama 3.1 8B & \textbf{0.472} & 0.542 & 0.615 & 0.699 & 0.792 \\
    % Llama 3.3 Instruct 70B & 0.398 & 0.513 & 0.597 & 0.665 & 0.734 \\
    MB Large 0.4B & 0.430 & 0.531 & 0.617 & 0.698 & 0.792 \\
    CodeBERT 0.1B & 0.327 & 0.476 & 0.564 & 0.659 & 0.762 \\
    Yi Coder 9B & 0.430 & 0.527 & 0.603 & 0.689 & 0.779 \\
    Yi Coder 1.5B & 0.460 & 0.542 & 0.619 & 0.698 & 0.791 \\
    \bottomrule
  \end{tabular}
\end{table}

\begin{table*}[bp]
  \caption{ONNX-to-text Encoding vs. PyTorch Text (NAS-Bench-201).}
  \label{tab:onnx_vs_pytorch}
  \begin{tabular}{llccccc}
    \toprule
    & & \multicolumn{5}{c}{Avg Kendall's Tau at Training Set Size} \\
    \cmidrule(lr){3-7}
    Encoding & Model & 14 & 55 & 220 & 879 & 3516 \\
    \midrule
    ONNX-to-text Encoding & CodeLlama Python 7B & 0.425 & 0.522 & 0.605 & 0.678 & 0.760 \\
    & ModernBERT Large & 0.416 & 0.510 & 0.578 & 0.638 & 0.721 \\
    \midrule
    PyTorch Text & CodeLlama Python 7B & \textbf{0.441} & \textbf{0.546} & \textbf{0.626*} & \textbf{0.707*} & \textbf{0.794*} \\
    & ModernBERT Large & 0.433 & 0.520 & 0.608* & 0.690* & 0.780* \\
    \bottomrule
    \multicolumn{7}{l}{\footnotesize * Statistically significantly better ($p < 0.05$) than the counterpart text encoding when using the same LM.}
  \end{tabular}
\end{table*}

\begin{table*}[bp]
  \caption{Derivation Tree String vs. PyTorch Text (einspace).}
  \label{tab:derivation_vs_pytorch}
  \begin{tabular}{llccccc}
    \toprule
    & & \multicolumn{5}{c}{Avg Kendall's Tau at Training Set Size} \\
    \cmidrule(lr){3-7}
    Encoding & Model & 14 & 55 & 220 & 879 & 2553 \\
    \midrule
    Derivation Tree String & CodeLlama Python 7B & 0.493 & 0.613 & 0.703 & 0.763 & 0.804 \\
    & ModernBERT Large & 0.489 & 0.594 & 0.683 & 0.751 & 0.801 \\
    \midrule
    PyTorch Text & CodeLlama Python 7B & \textbf{0.508} & \textbf{0.614} & \textbf{0.711} & 0.778* & 0.817* \\
    & ModernBERT Large & 0.475 & 0.599 & 0.705* & \textbf{0.780*} & \textbf{0.819*} \\
    \bottomrule
    \multicolumn{7}{l}{\footnotesize * Statistically significantly better ($p < 0.05$) than the counterpart text encoding when using the same LM.} \\
  \end{tabular}
\end{table*}

\subsubsection{Base configuration modifications}

Table~\ref{tab:kendall_results} presents the Kendall’s Tau performance across different training sample sizes for deviations from the base configuration. At low sample sizes ($N=14, 55$), applying PCA hurts performance. However, as the number of training samples increases ($N \ge 220$), PCA becomes critical. At $N=879$, the Base config ($0.703$) substantially outperforms No PCA ($0.645$). This suggests that while raw embeddings contain rich information, they are high-dimensional and prone to overfitting in the regressor head as data scales, while PCA effectively condenses the signal for the MLP. Meanwhile, 8-bit quantization of LM parameters improves performance at $N = 14$, but it is not superior at higher sample sizes. We also observe that for verbosity, the Inline mode and Helper Method mode perform comparably to the Base configuration (Excluded Helper Mode), demonstrating the LM's capability to infer operation semantics. We proceed with the Excluded Helper Mode as our standard because it is token-efficient. Additionally, using the Backbone add-on yields a consistent, albeit marginal, improvement over the Base configuration (e.g., $0.711$ vs. $0.703$ at $N=879$). This suggests the LM does utilize the global architectural context. However, given the small performance delta and large increase in input length (lowered token efficiency), we do not include the add-on in the COLE framework setup in following experiments.

\subsubsection{Loss Function Dynamics}
We then compared the Pairwise Hinge Loss against a standard Mean Squared Error (MSE) loss. As shown in Table~\ref{tab:mse_results}, minimizing MSE naturally results in lower absolute error. However, Table~\ref{tab:kendall_results} confirms that Pairwise Loss yields consistently higher Kendall’s Tau correlations ($0.703$ vs $0.691$ at $N=879$). Since the primary goal of a NAS surrogate is to correctly rank candidates for selection rather than predict their exact accuracy, we adopt the Pairwise Hinge Loss as the standard.

\subsubsection{LM Comparison}
We expanded our analysis to compare different LM families, sizes, and pre-training objectives. Table~\ref{tab:model_comparison} details the performance of various models under the Base configuration. The results underscore the importance of domain specificity. CodeLlama Python (both 7B and 34B) almost consistently lead the benchmarks. Interestingly, scaling from 7B to 34B offers only marginal gains (e.g., $0.703$ vs $0.711$ at $N=879$), and the smaller Yi Coder 1.5B model performs comparably (e.g., $0.698$ at $N = 879$). Decoder-only models (CodeLlama, Yi Coder) generally outperform Encoder-only models (ModernBERT, CodeBERT), but while ModernBERT Large is still comparable to others, CodeBERT lags significantly behind.

\subsection{Comparison to Other Text-Based Encodings}

We also evaluate whether PyTorch code representations outperform existing text-based encodings. We compare against ONNX-to-text encodings on NAS-Bench-201 and derivation tree strings on einspace.

\subsubsection{ONNX-to-text encodings}\label{onnx_experiment}
We compared our method against the ONNX-to-text encoding strategy \cite{qin2025onnx}, which serializes the architecture's computation graph into a text sequence.

We utilized the ONNX corpus provided by Qin et al. \cite{qin2025onnx}. As this corpus lacks direct identifiers to the original NAS-Bench-201 architectures, we aligned the datasets based on test accuracy (precision $10^{-7}$) of architectures. This process resulted in a perfectly matched subset of only 7,400 architectures out of the overall 15,625, since the accuracies reported by ONNX-bench do not perfectly align with those reported by the official NAS-Bench-201 API. To ensure a fair comparison, the stratified subsampled cross-validation for this experiment was restricted to this subset, with each of the 10 folds having size 740. As shown in Table~\ref{tab:onnx_vs_pytorch}, PyTorch text encoding consistently outperforms ONNX-to-text across both CodeLlama and ModernBERT. The gap is statistically significant at larger sample sizes ($N \ge 220$), suggesting that code aligns better with the pre-training distribution of these models than the ONNX-to-text encoding.

\subsubsection{Derivation Tree Strings}\label{devtree_experiment}
We extended our evaluation to einspace, a search space defined by a context-free grammar. We compare our PyTorch transpiler (explained in Section~\ref{einspaceTranspiler}) against the native derivation tree strings used by Qin et al. \cite{qin2025transferrable}.

Unlike the finite NAS-Bench-201, einspace is an open-ended space. To construct a representative corpus, we aggregated the search histories of 9 distinct NAS trials running Qin et al.’s codebase \cite{qin2025transferrable}. Each run explored einspace for an end-task of optimizing performance on CIFAR-10 classification. This yielded a diverse dataset of 2,837 unique architectures possessing both a derivation tree string and a corresponding PyTorch code representation. The same stratified subsampled cross-validation protocol is again used, with each fold having an approximate size of 283. The architectures’ validation accuracies range between 9.28\% and 82.5\%.

Table~\ref{tab:derivation_vs_pytorch} demonstrates that PyTorch encoding outperforms the derivation tree string representation, particularly for the CodeLlama model. At $N=879$, PyTorch encoding achieves a $\tau$ of 0.778 compared to 0.763 for derivation trees. This shows that even for grammar-based spaces, translating the abstract tree into standard Python code allows the LM to leverage its pre-trained knowledge more effectively.

\begin{figure}[t]
  \centering
  \includegraphics[width=\linewidth]{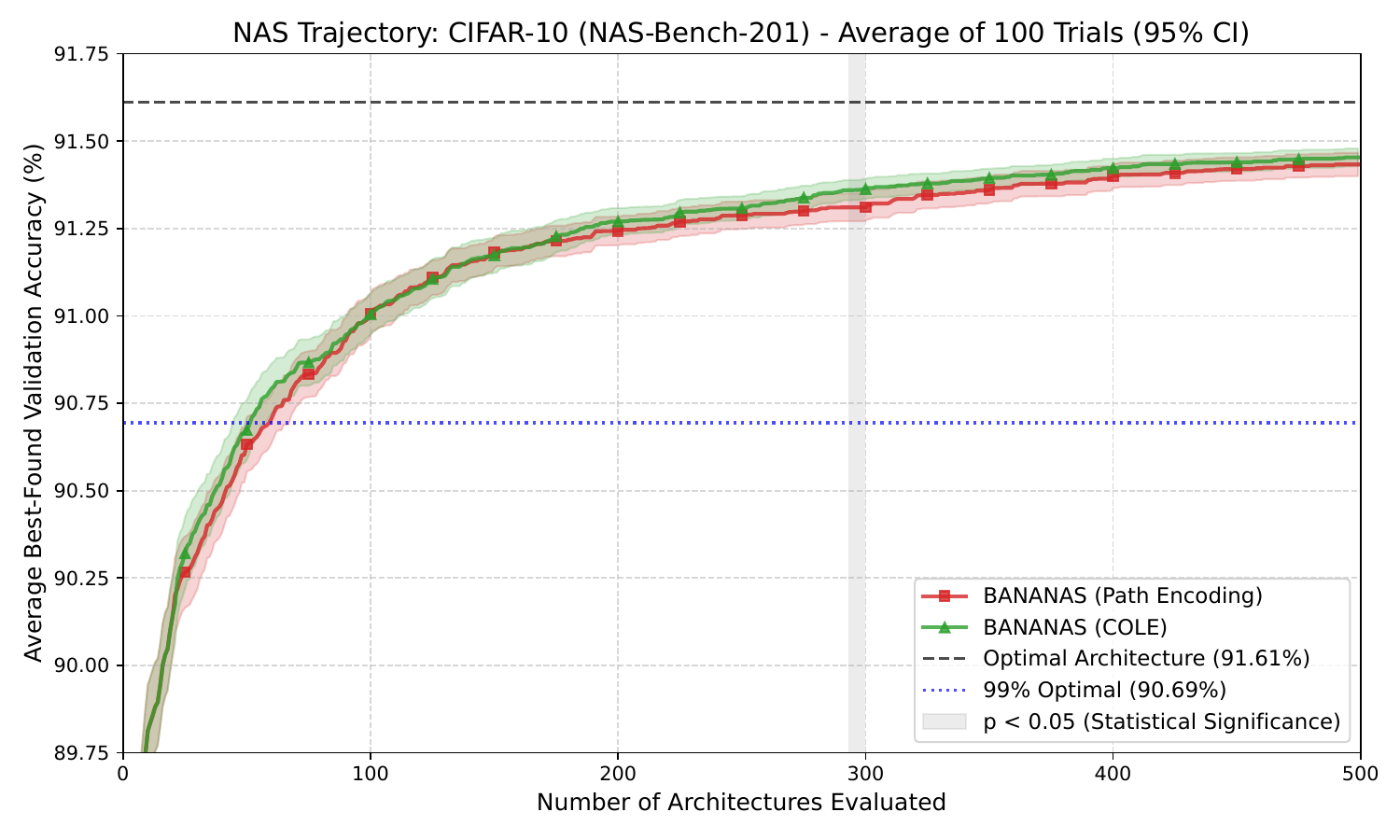} 
  \includegraphics[width=\linewidth]{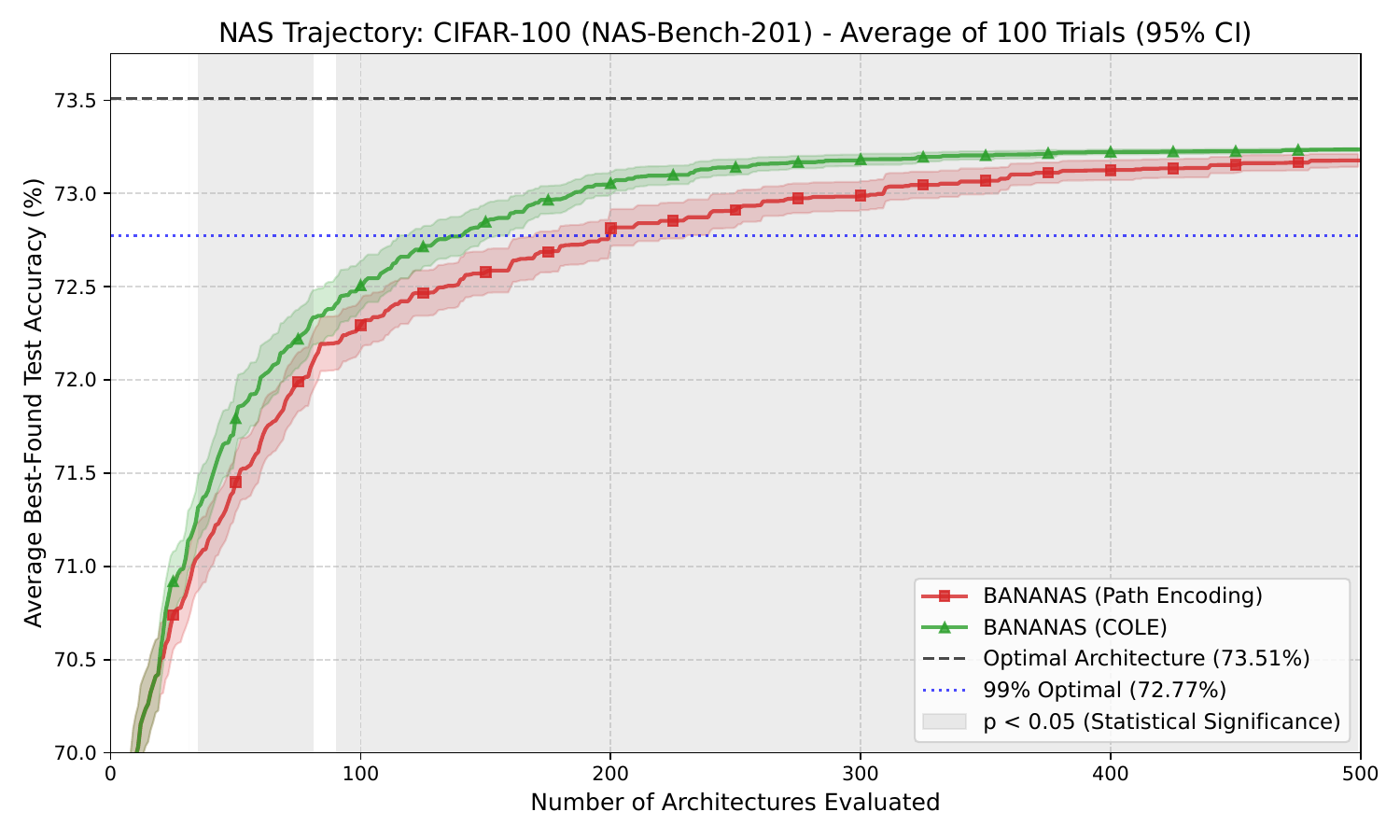} 
  \includegraphics[width=\linewidth]{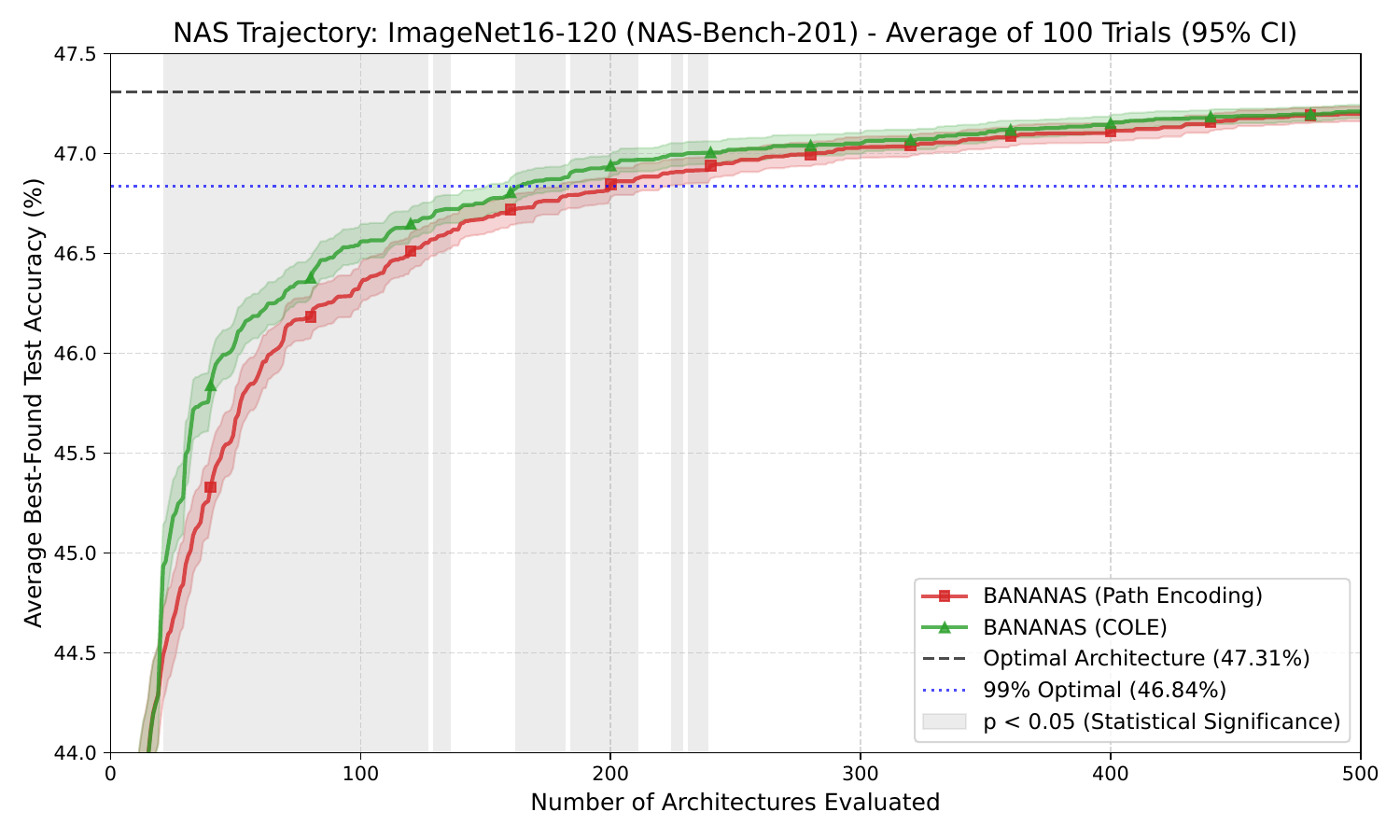} 
  \caption{NAS trajectory on CIFAR-10 (top), CIFAR-100 (middle), and ImageNet16-120 (bottom) in NAS-Bench-201 (100 trials each). A line graph shows Average Best-Found (Test or Validation) Accuracy vs. Number of Architectures Evaluated. The comparison between COLE (green) vs path encodings (red) with the BANANAS algorithm are shown. Colored shading indicates a 95\% confidence interval, and vertical gray shading indicates stages at which COLE's performance is statistically significantly different.}
  \Description{}
  \label{fig:trajectory}
\end{figure}
\subsection{Downstream Search Efficiency}
To validate the utility of our embeddings for NAS, we integrated COLE as a custom representation for architectures within the NASLib framework \cite{white2021powerful}. We conducted NAS in the NAS-Bench-201 search space to find architectures optimal for three classification end-tasks: CIFAR-10 validation accuracy, CIFAR-100 test accuracy, and ImageNet16-120 test accuracy. The control baseline was the standard BANANAS algorithm, which uses path encodings to represent architectures. We replaced these encodings with COLE while maintaining the same surrogate predictor architecture, a 3-layer MLP with a hidden layer size of 128, with implementation already provided in NASLib. We executed 100 independent trials with random initialization for each method \& end-task, setting a budget of 500 ground-truth architecture evaluations. The first 20 evaluations are guided by the regularized evolution algorithm, before BANANAS is activated, acting as a population initialization step. The surrogate retrains every 10 evaluations. We report the average of the best-found architecture's accuracy relative to the number of evaluations performed.

The search trajectories (Figure~\ref{fig:trajectory}, Table~\ref{tab:evaluations_comparison}) demonstrate that replacing structural encodings with COLE enhances search efficiency. The CIFAR-100 performance gap is statistically significant ($p < 0.05$) through almost the entire search, while the ImageNet16-120 performance gap is statistically significant in the early stages of search. Using COLE, BANANAS discovered near-optimal architectures that are within 1\% of the top CIFAR-10 performer's accuracy (which is 91.61\%) in 52 evaluations, compared to 60 evaluations required when using path encodings (i.e., a 13\% reduction). Similarly, for CIFAR-100 (top accuracy 73.51\%), COLE required 132 evaluations, compared to 200 required by path encodings (i.e., a 34\% reduction). For ImageNet16-120 (top accuracy 47.31\%), COLE required 164 evaluations compared to 200 required by path encodings (i.e., a 18\% reduction). These reductions are crucial since true model evaluations are computationally prohibitive during real-world NAS. Although BANANAS is Bayesian Optimization-based, COLE is purely a feature extractor and thus can enhance surrogate models in NAS algorithms that are necessarily evolutionary too.

\newcolumntype{C}{>{\centering\arraybackslash}p{1.9cm}}
\begin{table*}[t]
  \centering
  \caption{Average best-found accuracy ($\pm$ standard deviation) values from the NAS trajectories shown in Figure~\ref{fig:trajectory} at different evaluation budgets. The accuracy metric reported for the CIFAR-10 end-task is validation accuracy, while test accuracy is reported for CIFAR-100 and ImageNet16-120.}
  \label{tab:evaluations_comparison}
  \begin{tabular}{llCCCC}
    \toprule
    & & \multicolumn{4}{c}{Avg Best-Found Accuracy (\%) $\pm$ Std at Evaluation Budget} \\
    \cmidrule(lr){3-6}
    End-task & Encoding & 60 & 120 & 250 & 500 \\
    \midrule
    CIFAR-10 & COLE & \textbf{90.78} $\pm$0.36 & 91.08 $\pm$0.28 & \textbf{91.31} $\pm$0.17 & \textbf{91.45} $\pm$0.13 \\
    & Path & 90.69 $\pm$0.36 & 91.08 $\pm$0.26 & 91.29 $\pm$0.20 & 91.43 $\pm$0.17 \\
    \midrule
    CIFAR-100 & COLE & \textbf{71.95*} $\pm$0.87 & \textbf{72.66*} $\pm$0.58 & \textbf{73.14*} $\pm$0.21 & \textbf{73.24*} $\pm$0.04 \\
    & Path & 71.61 $\pm$0.83 & 72.42 $\pm$0.61 & 72.91 $\pm$0.47 & 73.18 $\pm$0.17 \\
    \midrule
    ImageNet16-120 & COLE & \textbf{46.19}* $\pm$0.11 & \textbf{46.63}* $\pm$0.08 & \textbf{47.02} $\pm$0.06 & \textbf{47.21} $\pm$0.03 \\
    & Path & 45.89 $\pm$0.13 & 46.49 $\pm$0.09 & 46.95 $\pm$0.06 & 47.20 $\pm$0.04 \\
    \bottomrule
    \multicolumn{6}{l}{\footnotesize * Statistically significantly better ($p < 0.05$) than the counterpart encoding when optimizing for the same end-task.} \\
  \end{tabular}
\end{table*}

\section{Discussion}

This study demonstrates that frozen LMs act as effective, zero-shot architecture embedders for Neural Architecture Search. Our results position COLE as a distinct and improved alternative to structural encoding techniques (like path encodings) and existing text-based encodings. By using raw code as the input modality, we bypass the need for expensive domain-specific fine-tuning while achieving strong predictive performance in ``cold-start'' scenarios. 
\subsection{Verbosity and Token Efficiency}
Our empirical analysis reveals that the COLE framework is robust to input verbosity. We observed negligible performance differences between providing full operation implementations (Inline and Helper Method modes) versus simply invoking function signatures (Excluded Helper mode). This suggests that the LM's pre-trained inductive bias allows it to infer operation semantics (e.g., knowing what a \code{ReLU\_Conv2d\_BatchNorm} block implies) without needing to process the boilerplate implementation code. 

\subsection{Comparison to Structural Encodings} Traditional methods like adjacency matrices or path encodings (used in BANANAS) are computationally efficient. Contrarily, extracting embeddings from a 7B parameter model is orders of magnitude more expensive than calculating a path string. We acknowledge that recent domain-specific predictors such as FLAN \cite{akhauri2024encodings} or HyperNAS \cite{lv2025hypernas}, which utilize similar structural encodings, often achieve Kendall’s Tau correlations higher than COLE. But, these encodings are semantically incomplete. They often treat operations as discrete labels or entries, so the predictor model must learn the meaning and interaction of operations from scratch. These encodings also struggle to capture parametric details such as kernel sizes, padding, or activation types without complex, manual search-space-specific engineering. This makes it difficult to apply across different types of architectures or use cases. In contrast, as a code-based approach, COLE resolves problems of both capturing semantics and flexible applicability. Code naturally defines both the topology (via variable flow: \code{z = x + y}) and the hyperparameters (\code{kernel\_size=3}) in a unified format; embedding using an LM helps these components become interpretable numerical data. Furthermore, since code is a universal descriptor, this advantage (and the broader methodology) can apply immediately to new search spaces without the need to engineer a new domain-specific matrix encoding scheme or train a complex GNN predictor from scratch. In summary, COLE's primary strength is ease of use, functioning as a generalizable alternative that helps in ``cold-start'' circumstances. 

\subsubsection{Additional Notes on Efficiency}
Our analysis of model sizes challenges the assumption that massive models are required for effective embeddings. The Yi Coder 1.5B and ModernBERT Large (0.4B) models achieved comparable performance with the 7B and 8B parameter models.

\subsection{Comparison to Other Text-based Encodings} Recent LM-based approaches attempt to unify encodings by serializing architectures into ONNX-to-text encodings \cite{qin2025onnx} or derivation tree strings \cite{qin2025transferrable}. While generalizable, they do not align with the LMs’ pre-training data. LMs' pre-training often consists of programming languages and code (e.g., Python) as input. By feeding an LM novel text-based encodings, prior works force the model to interpret an underrepresented format. In contrast, our method matches the distribution of data seen during the LM's pre-training, allowing us to improve performance when bypassing the fine-tuning methodology and leverage the model's existing world knowledge of neural networks. In fact, while Qin et al. \cite{qin2025transferrable} achieve strong results after fine-tuning, our results show that COLE enables superior ranking correlation out-of-the-box with a lower quantity of training samples required. When operating in einspace, COLE achieves a validation $\tau \approx 0.71$ at just $N = 220$ training samples, while Qin et al. \cite{qin2025transferrable} only achieve $\tau = 0.628$ after training on approximately 6.5k architectures. This makes COLE more practical for resource-constrained NAS.

\subsection{Limitations and Future Directions}
Despite the promising results of using COLE, we acknowledge several limitations in our current study. Our downstream search evaluation was restricted to NAS-Bench-201; validating the efficiency gains on larger, more diverse search spaces like DARTS \cite{liu2018darts} remains necessary. Our comparative analysis against other text-based encodings was limited to NAS-Bench-201 and einspace. Expanding this validation to other search spaces would strengthen the generalizability of our findings regarding the superiority of native code.
Future work can focus on optimizing the embedding extraction pipeline, including advanced pooling strategies or architecture of the regression head. Furthermore, while we focused on frozen models to minimize cost, fine-tuning LMs on the specific syntax of neural architectures still represents a promising avenue to boost performance. Additionally, the unified nature of code representations naturally facilitates joint architecture and hyperparameter search, enabling the simultaneous optimization of topological structure and training scheme (e.g., optimizers and learning rate schedulers). Finally, extending COLE to multi-objective NAS, such as hardware awareness, offers significant potential.

\begin{acks}
We used Github Copilot for assistance in creating programs to execute comparison experiments. All logic was verified and corrected by the authors. 

Our research was also supported by the AI Makerspace of the College of Engineering (RRID:SCR\_028058), provided by the Partnership for an Advanced Computing Environment (PACE) at the Georgia Institute of Technology (RRID:SCR\_027619).
\end{acks}

% %% The next two lines define the bibliography style to be used, and
% %% the bibliography file.
\bibliographystyle{ACM-Reference-Format}
\bibliography{samples/references}

\newpage
\appendix

\section{Code Verbosity: Context Add-ons}\label{addon_appendix}

As discussed in Section~\ref{code_for_nb201}, we evaluated whether providing LMs with broader architectural and training context improves embedding quality.

\begin{figure}[h]
  \begin{lstlisting}[language=Python]
class Network(nn.Module):
  def __init__(self, channels, N, genotype, num_classes):
    super(Network, self).__init__()
    self.C = channels
    self.N = N
    self.stem = nn.Sequential(
      nn.Conv2d(3, self.C, kernel_size=3, padding=1, bias=False),
      nn.BatchNorm2d(self.C)
    )
    
    layer_channels = [self.C] * N + [self.C * 2] + [self.C * 2] * N + [self.C * 4] + [self.C * 4] * N
    layer_reductions = [False] * N + [True] + [False] * N + [True] + [False] * N

    C_prev = self.C
    self.cells = nn.ModuleList()
    for index, (C_curr, reduction) in enumerate(
      zip(layer_channels, layer_reductions)
    ):
      if reduction:
        cell = ResNetBasicblock(C_prev, C_curr, 2, True)
      else:
        cell = Cell(C_curr)
      self.cells.append(cell)
      C_prev = C_curr
    
    self._Layer = len(self.cells)
    self.lastact = nn.Sequential(nn.BatchNorm2d(C_prev), nn.ReLU(inplace=True))
    self.global_pooling = nn.AdaptiveAvgPool2d(1)
    self.classifier = nn.Linear(C_prev, num_classes)

  def get_training_config(self):
    optimizer = torch.optim.SGD(
      self.parameters(), lr=0.1, momentum=0.9,
      weight_decay=5e-4, nesterov=True
    )
    scheduler = torch.optim.lr_scheduler.CosineAnnealingLR(optimizer, T_max=200, eta_min=0)
    config = {
      'optimizer': optimizer, 'scheduler': scheduler,
      'batch_size': 256, 'epochs': 200
    }
    return config

  def forward(self, inputs):
    feature = self.stem(inputs)
    for i, cell in enumerate(self.cells):
      feature = cell(feature)
    out = self.lastact(feature)
    out = self.global_pooling(out)
    out = out.view(out.size(0), -1)
    logits = self.classifier(out)
    return out, logits
  \end{lstlisting}
  
  \caption{An excerpt of the Backbone add-on optionally used in COLE. The implementations for the referenced \code{ResNetBasicblock} and its dependency, the \code{ReLUConvBN} operation, are not shown here but are included in the PyTorch code string when the Backbone add-on is used.}
  \label{fig:backbone_addon}
\end{figure}

Figure \ref{fig:backbone_addon} displays an excerpt of the Backbone add-on. We used syntax of the official macroskeleton implementation from a NAS-Bench-201 codebase, making minimal modifications such as including the training recipe. Figure \ref{fig:comment_addon} displays the Comment add-on, which condenses this same topological and training information into a natural language docstring. When used, these add-ons are prepended to the base PyTorch \code{Cell} class definition to build the code representation.

\begin{figure}[h]
  \begin{lstlisting}[numbers=none, frame=single]
Task: CIFAR-10 image classification (10 classes, 32x32 RGB images).

This Cell is one building block within a larger neural network.
Full architecture:
- Stem layer: Conv2d(3 channels -> 16 channels, 3x3 kernel) + BatchNorm2d.
- Main head: stacks 15 copies of the Cell into a sequence. 1 ResNetBasicblock layer is inserted every 5 Cells (total 2).
- Final layers: BatchNorm2d + ReLU + Global Average Pooling + Linear layer to 10 classes.

Helpers:
- ReLUConvBN: Sequential ReLU -> Conv -> BatchNorm (pre-activation)
- ResNetBasicblock: Residual block with 2 ReLUConvBN plus 1 skip connection with input downsampling

Training Details: SGD optimizer with momentum=0.9, weight_decay=5e-4, initial learning_rate=0.1 
with cosine annealing schedule over 200 epochs, batch_size=256, plus standard data augmentation.
  \end{lstlisting}
  
  \caption{The Comment add-on optionally used in COLE. A summary of the cell, macroskeleton (Backbone), task, and training details is provided.}
  \label{fig:comment_addon}
\end{figure}

\section{einspace Representations}\label{einspace_appendix}
As discussed in Section~\ref{einspaceTranspiler}, we translated einspace architectures into code from their derivation tree representations. Figure ~\ref{fig:devtree_string} and Figure~\ref{fig:einspace_code_example} show the derivation tree string and corresponding PyTorch code of a particular sample architecture, respectively.

\begin{figure}[h]
  \begin{lstlisting}[numbers=none, frame=single]
branching(4)[
  clone(4){'out_feature_shape': [3, 32, 32]},
  sequential[
    routing[
      im2col(3,2,1){'out_feature_shape': [256, 27]},
      computation[linear(32){'out_feature_shape': [256, 32]}],
      identity{'out_feature_shape': [256, 32]}
    ],
    computation[linear(16){'out_feature_shape': [256, 16]}]
  ],
  cat(4,1){'out_feature_shape': [1024, 16]}
]
  \end{lstlisting}
  
  \caption{The derivation tree string representation for a sample einspace architecture.}
  \label{fig:devtree_string}
\end{figure}

\begin{figure}[h]
  \begin{lstlisting}[numbers=none, frame=single]
class Network(nn.Module):
  def __init__(self):
    super(Network, self).__init__()
    self.clone_0 = CloneTensor(num_clones=4)
    self.im2col_0 = Im2Col(input_shape=[1, 3, 32, 32], kernel_size=3, stride=2, padding=1)
    self.linear_0 = nn.Linear(27, 32)
    self.computation_0 = ComputationModule(computation_fn=self.linear_0)
    self.identity_0 = nn.Identity()
    self.routing_0 = RoutingModule(
      prerouting_fn=self.im2col_0,
      inner_fn=self.computation_0,
      postrouting_fn=self.identity_0
    )
    self.linear_1 = nn.Linear(32, 16)
    self.computation_1 = ComputationModule(computation_fn=self.linear_1)
    self.sequential_0 = SequentialModule(
      first_fn=self.routing_0,
      second_fn=self.computation_1
    )
    self.cat_0 = CatTensors(dim=1)
    self.branching_0 = BranchingModule(
      branching_fn=self.clone_0,
      inner_fn=nn.ModuleList([self.sequential_0]),
      aggregation_fn=self.cat_0
    )
  def forward(self, x):
    return self.branching_0(x)
  \end{lstlisting}
  
  \caption{The PyTorch code representation for the einspace architecture shown in Figure~\ref{fig:devtree_string}.}
  \label{fig:einspace_code_example}
\end{figure}

\section{Encoding Visualization}\label{tsne_viz}
To better understand how different representations organize the search space, we visualize the embeddings of architectures from NAS-Bench-201 and einspace. 

\subsection{NAS-Bench-201 (Structural and COLE)}
\begin{figure*}
  \centering
  \includegraphics[width=\linewidth]{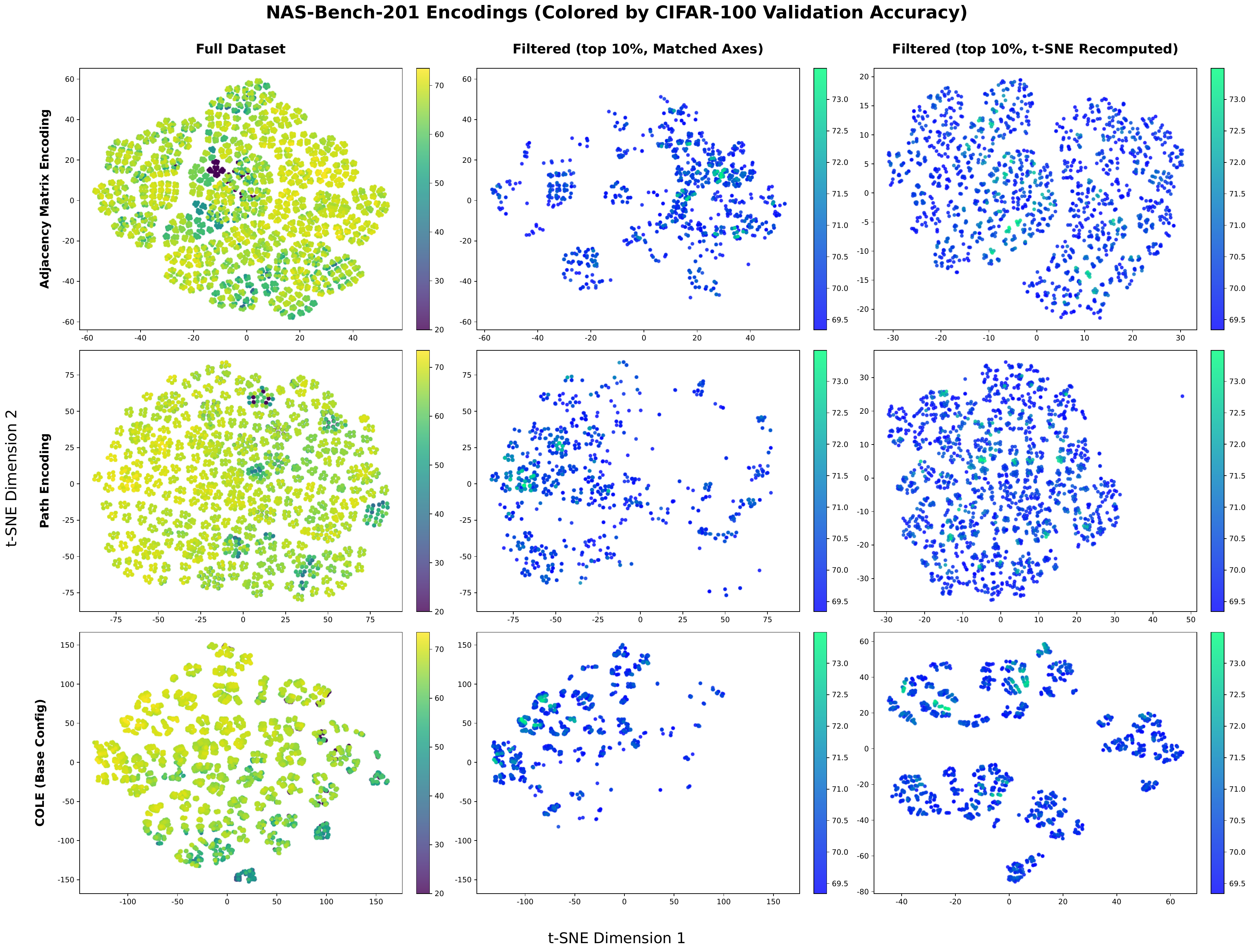} 
  \caption{t-SNE visualizations comparing structural encodings with COLE on NAS-Bench-201. The left column shows the full search space. The middle column filters for the top 10\% of architectures and plots using the original t-SNE projection. The right column recomputes the t-SNE projection on the top 10\%. Left column: architectures having accuracy $\leq$20\% are painted the same color.}
  \Description{}
  \label{fig:tsne_nb201}
\end{figure*}

Figure~\ref{fig:tsne_nb201} compares COLE (generated using CodeLlama Python 7B) against two standard structural encodings: adjacency matrices and path encodings. We use t-SNE to project the entire NAS-Bench-201 search space into 2D, as shown in the leftmost column of plots in Figure~\ref{fig:tsne_nb201}. Color-coding indicates architectures' CIFAR-100 validation accuracy. To highlight the distribution of the best models, the middle column filters to isolate the top 10\% of architectures and displays them within the same t-SNE projection, i.e. without recomputing the t-SNE components. The right column shows plots after indeed recomputing t-SNE solely on this top-10\% subset.

In the structural encoding spaces, high-performing architectures cluster moderately but are still scattered throughout the t-SNE projection. By contrast, COLE appears to better split high- and low-performing architectures in the t-SNE projection. A higher concentration of higher-performing architectures tends to appear on one particular side of the projection. This organization is especially apparent in the middle column of Figure~\ref{fig:tsne_nb201}, when isolating the top 10\%. While the structural encodings also seem to form some separation of high- and low-performers, the effect appears stronger under COLE. This suggests that our code-based embeddings inherently capture performance-relevant semantics better than structural encodings.

\subsection{NAS-Bench-201 (ONNX)}
\begin{figure*}
  \centering
  \includegraphics[width=\linewidth]{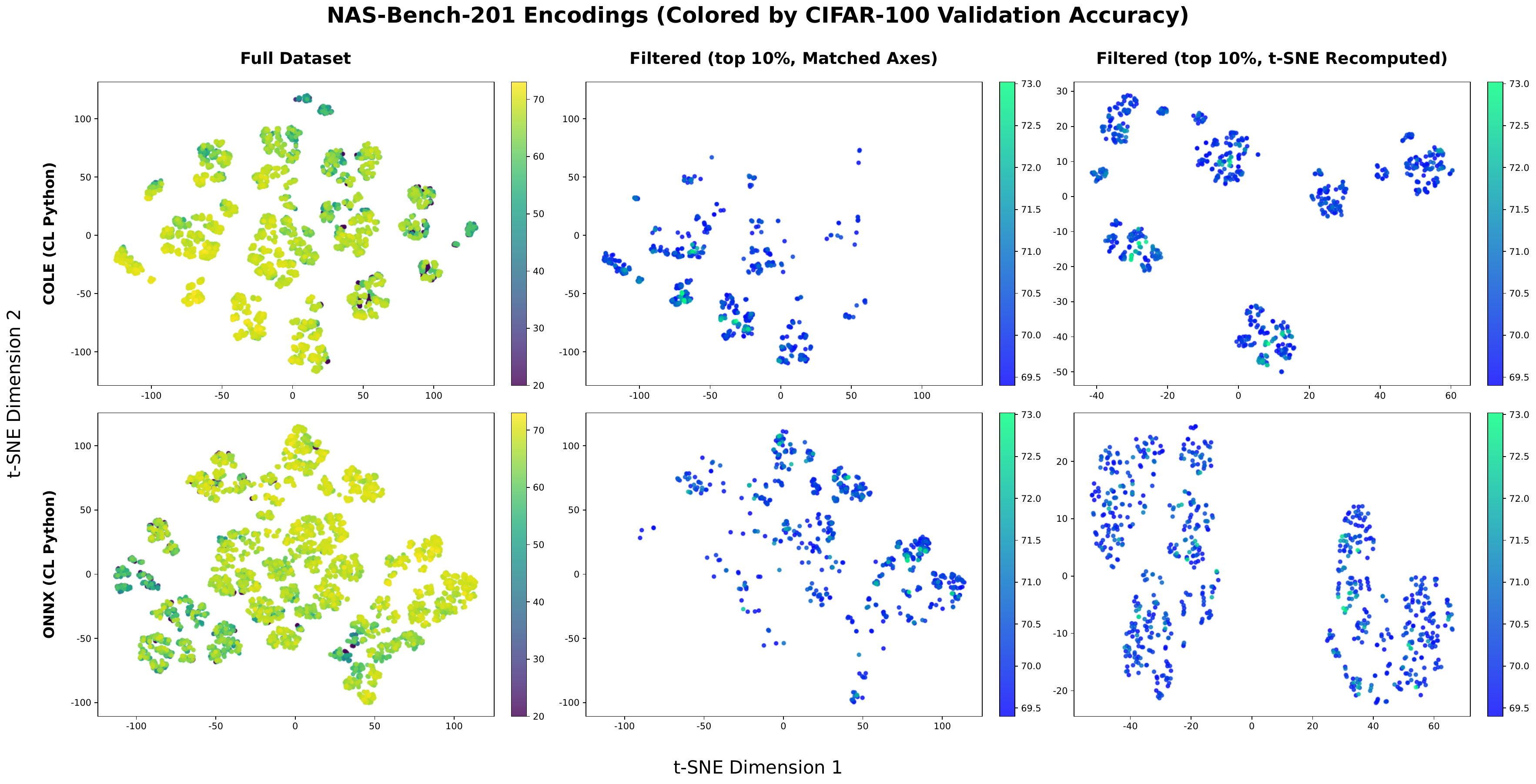} 
  \caption{t-SNE visualizations comparing ONNX-to-text encodings with COLE on a 7,400-architecture subset of NAS-Bench-201. The layout follows Figure~\ref{fig:tsne_nb201}, filtering and recomputing the projection for the top 10\% of architectures. Left column: architectures having accuracy $\leq$20\% are painted the same color. CL: CodeLlama.}
  \Description{}
  \label{fig:tsne_onnx}
\end{figure*}

Following the experiments in Section~\ref{onnx_experiment}, Figure~\ref{fig:tsne_onnx} compares COLE against ONNX-to-text encodings. Because the ONNX dataset only successfully aligned with a subset of 7,400 NAS-Bench-201 architectures, these t-SNE projections are computed exclusively on that exact subset.

\subsection{einspace (Derivation Tree)}
\begin{figure*}
  \centering
  \includegraphics[width=\linewidth]{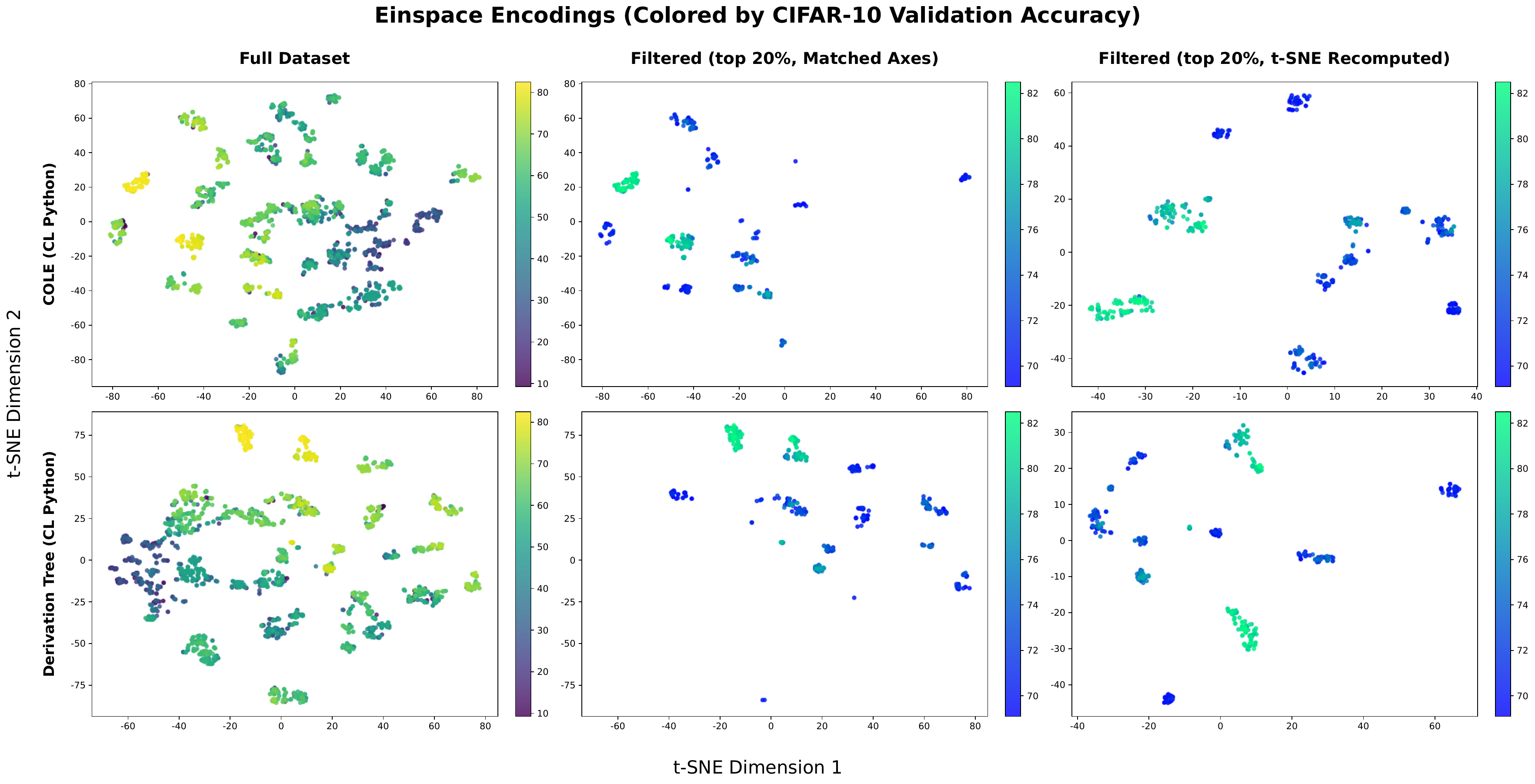} 
  \caption{t-SNE visualizations comparing derivation tree strings and COLE on a 2,837-architecture corpus. The layout follows Figure~\ref{fig:tsne_nb201}, filtering and recomputing the projection for the top 20\% of architectures. CL: CodeLlama.}
  \Description{}
  \label{fig:tsne_einspace}
\end{figure*}

Similarly, Figure~\ref{fig:tsne_einspace} visualizes the einspace search space, corresponding to the experiments in Section~\ref{devtree_experiment}. These plots compare COLE against derivation tree strings across the set of 2,837 architectures discovered during the einspace search trials.

\end{document}